\documentclass{article}
\usepackage{arxiv}
\usepackage[utf8]{inputenc}
\usepackage{array}
\usepackage[hidelinks]{hyperref}
\usepackage[caption=false]{subfig}
\usepackage[pdftex]{graphicx}
\usepackage[mode=buildnew]{standalone}
\usepackage{tikz}
\usepackage{booktabs}
\usepackage{multirow}
\usepackage{siunitx}

\newcommand{\samplesize}{2.2cm}
\newcommand{\samplerise}{-1cm}
\newcommand*\rot{\rotatebox{90}}
\title{Sensing Anomalies as Potential Hazards: Datasets and Benchmarks}

\author{Dario Mantegazza, Carlos Redondo, Fran Espada, Luca M. Gambardella, Alessandro Giusti and J\'er\^ome Guzzi 
\thanks{Dario Mantegazza, Luca M. Gambardella, Alessandro Giusti, and J\'er\^ome Guzzi are with the Dalle Molle Institute for Artificial Intelligence (IDSIA), USI-SUPSI, Lugano, Switzerland. Carlos Redondo and Fran Espada are with Hovering Solutions Ltd, Madrid, Spain. 
This work is supported by the Swiss National Science Foundation (SNSF) through the NCCR Robotics and by the European Commission through the Horizon 2020 project 1-SWARM, grant ID 871743.}
}
\date{September 2022}

\begin{document}

\maketitle
\thispagestyle{empty}
\pagestyle{empty}

\begin{abstract}
We consider the problem of detecting, in the visual sensing data stream of an autonomous mobile robot, semantic patterns that are unusual (i.e., anomalous) with respect to the robot's previous experience in similar environments.  These anomalies might indicate unforeseen hazards and, in scenarios where failure is costly, can be used to trigger an avoidance behavior.  We contribute three novel image-based datasets acquired in robot exploration scenarios, comprising a total of more than 200k labeled frames, spanning various types of anomalies.  On these datasets, we study the performance of an anomaly detection approach based on autoencoders operating at different scales.

\keywords{Visual Anomaly Detection \and Dataset for Robotic Vision \and Deep Learning for Visual Perception \and Robotic Perception}
\end{abstract}

\section*{Supplementary Material}
Code, video, and data available at\\ \url{{https://github.com/idsia-robotics/hazard-detection}}.

\section{Introduction}
\label{sec:intro}


Many emerging applications involve a robot operating autonomously in an unknown environment; the environment may include hazards, i.e., locations that might disrupt the robot's operation, possibly causing it to crash, get stuck, and more generally fail its mission.
Robots are usually capable to perceive hazards that are expected during system development and therefore can be explicitly accounted for when designing the perception subsystem.  For example, ground robots can typically perceive and avoid obstacles or uneven ground.

In this paper, we study how to provide robots with a different capability: detecting \emph{unexpected} hazards, potentially very rare, that were not explicitly considered during system design.  Because we don't have any model of how these hazards appear, we consider anything that is novel or unusual as a potential hazard to be avoided.


Animals and humans exhibit this exact behavior~\cite{wilson1994shyness}, known as \emph{neophobia}~\cite{moretti2015influence}: ``the avoidance of an object or other aspect of the environment solely because it has never been experienced and is dissimilar from what has been experienced in the individual's past''~\cite{stowe2006effects}.  We argue that autonomous robots could benefit from implementing neophobia, in particular whenever the potential failure bears a much higher cost than the avoidance behavior. 
Thus, for example, for a ground robot it makes sense to avoid unusual-looking ground~\cite{wellhausen2020safe} when a slightly longer path on familiar ground is available; or a planetary rover might immediately stop a planned trajectory if something looks odd, waiting for further instructions from the ground control.

Our experiments are motivated by a similar real-world use case in which a quadrotor equipped with sophisticated sensing and control traverses underground tunnels for inspection of aqueduct systems. During the flights, that might span several kilometers, the robot is fully autonomous since it has no connectivity to the operators; they wait for the robot to either reach the predetermined exit point or --- in case the robot decides to abort the mission --- backtrack to the entry.  In this context, a crash bears the cost of the lost hardware and human effort, but most importantly the lost information concerning the hazard that determined the failure, that remains unknown.
It then makes sense to react to unexpected sensing data by aborting the mission early and returning to the entry point;\footnote{Similarly, retention of information following encounters with novel predators is one of the recognized evolutionary advantages of neophobic animals~\cite{mitchell2016living}.}
operators can then analyze the reported anomaly: in case it is not a genuine hazard, the system can be instructed to ignore it in the current and future missions, and restart the exploration.

\begin{figure}[t]
    \centering
    \includegraphics[width=0.5\textwidth]{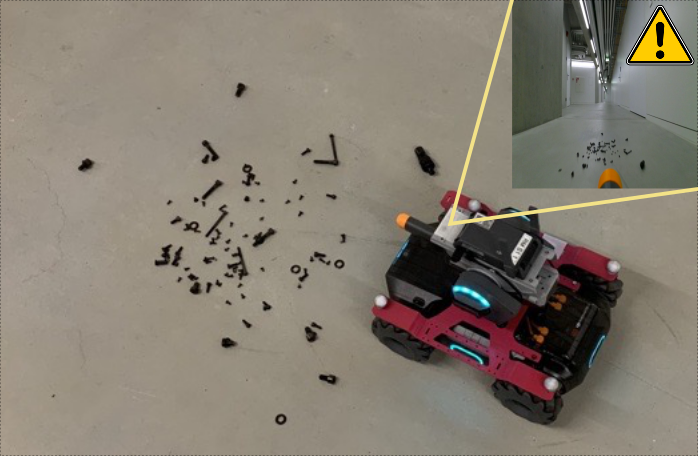}
    \caption{A Robomaster detects an anomaly in the camera frame: cautiousness is required.}
    \label{fig:RM}
\end{figure}

After reviewing related work (Section~\ref{sec:related}), we introduce in Section~\ref{sec:dataset} our \textbf{main contribution}: three image-based datasets (one simulated, two real-world) from indoor environment exploration tasks using ground or flying robots; each dataset is split into training (only normal frames) and testing sets; testing frames are labeled as normal or anomalous, representing hazards that are meaningful in the considered scenarios, including sensing issues and localized or global environmental hazards.  In Section~\ref{sec:experimental_setup}, we describe an anomaly detection approach based on autoencoders, and in Section~\ref{sec:experimental_results} we report and discuss extensive experimental results on these datasets, specifically exploring the impact of image sampling and preprocessing strategies on the ability to detect hazards at different scales.


\section{Related Work}
\label{sec:related}

\subsection{Anomaly Detection Methods}
Anomaly Detection (AD) is a widely researched topic in Machine Learning; general definitions of anomalies are necessarily vague: e.g., ``an observation that deviates considerably from some concept of normality''~\cite{ruff2021unifying}, or ``patterns in data that do not conform to expected behavior''~\cite{chandola2009anomaly}.  When operating on high-dimensional inputs, such as images, the problem often consists in finding \emph{high-level} anomalies~\cite{ruff2021unifying} that pertain to the data semantics, and therefore imply some level of understanding of the inputs.  Methods based on deep learning have been successful in high-level anomaly detection, in various fields, including medical imaging~\cite{schlegl2017unsupervised}, industrial manufacturing~\cite{scime2018multi,haselmann2018anomaly}, surveillance~\cite{chakravarty2007anomaly}, robot navigation~\cite{wellhausen2020safe}, fault detection~\cite{khalastchi2015online}, intrusion detection~\cite{birnbaum2015unmanned} and agriculture~\cite{christiansen2016deepanomaly}.


A widespread class of approaches for anomaly detection on images, which we adopt in this paper as a benchmark, is based on undercomplete autoencoders ~\cite{kramer1992autoassociative,cho2014learning}: neural reconstruction models that take the image as input and are trained to reproduce it as their output (e.g., using a Mean Absolute Error loss), while constraining the number of nodes in one of the hidden layers (the \emph{bottleneck}); this limits the amount of information that can flow through the network, and prevents the autoencoder from learning to simply copy the input to the output.  To minimize the loss on a large dataset of normal (i.e., non-anomalous) samples, the model has to learn to compress the inputs to a low-dimensional representation that captures their high-level information content.  When tasked to encode and decode an anomalous sample, i.e., a sample from a different distribution than the training set, one expects that the model will be unable to reconstruct it correctly.  Measuring the reconstruction error for a sample, therefore, yields an indication of the sample's anomaly.  Variational Autoencoders~\cite{kingma2013auto} and Generative Adversarial Networks (GAN)~\cite{goodfellow2014generative} can also be used for Anomaly Detection tasks, by training them to map vectors sampled from a predefined distribution (i.e., Gaussian or uniform) to the distribution of normal training samples. Flow-based generative models~\cite{dinh2016density} explicitly learn the probability density function of the input data using Normalizing Flows~\cite{kobyzev2020normalizing}.


One-Class Classifiers, such as Deep SVDD~\cite{ruff2018deep} and deep OC-SVM~\cite{erfani2016high}, can also be used as anomaly detectors; these methods define a decision boundary around the training instances in their respective latent spaces.


\subsection{Anomaly Detection on Images}
In recent work, Sabokrou et al.~\cite{sabokrou2018adversarially} propose a new adversarial approach using an autoencoder as a reconstructor, feeding a standard CNN classifier as a discriminator, trained adversarially.  During inference, the reconstructor is expected to enhance the inlier samples while distorting the outliers; the discriminator's output is used to indicate anomalies.

Sarafijanovic introduces~\cite{sarafijanovic2019fast} an Inception-like autoencoder for the task of anomaly detection on images. The proposed method uses different convolution layers with different filter sizes all at the same level, mimicking the Inception approach~\cite{szegedy2015going}. The proposed model works in two phases; first, it trains the autoencoder only on normal images, then, instead of the autoencoder reproduction error, it measures the 
distance over the pooled bottleneck's output, which keeps the memory and computation needs at a minimum. The authors test their solution over some classical computer vision datasets: MNIST~\cite{deng2012mnist}, Fashion MNIST~\cite{xiao2017fashion}, CIFAR10, and CIFAR100~\cite{krizhevsky2009learning}.

\subsection{Application to Robotics}
\paragraph{Using Low-Dimensional Data}
Historically, anomaly detection in robotics has focused on using low-dimensional data streams from exteroceptive or proprioceptive sensors. The data, potentially high-frequency, is used in combination with hand-crafted feature selection, Machine Learning, and, recently, Deep Learning models. Khalastchi et al.~\cite{khalastchi2011online,khalastchi2015online} use simple metrics such as Mahalanobis Distance to solve the task of online anomaly detection for unmanned vehicles; Sakurada et al.~\cite{sakurada2014anomaly} compare autoencoders to PCA and kPCA using spacecraft telemetry data. Birnbaum~\cite{birnbaum2015unmanned}, builds a nominal behavior profile of Unmanned Aerial Vehicle (UAV) sensor readings, flight plans, and state and uses it to detect anomalies in flight data coming from real UAVs. The anomalies vary from cyber-attacks and  sensor faults to structural failures.
Park et al. tackle the problem of detecting anomalies in robot-assisted feeding, in an early work the authors use Hidden Markov Models on hand-crafted features~\cite{park2016multimodal}; in a second paper, they solve the same task using a combination of Variational Autoencoders and LSTM networks~\cite{park2018multimodal}.
\paragraph{Using high-dimensional data}
An early approach~\cite{chakravarty2007anomaly} to anomaly detection on high-dimensional data relies on image matching algorithms for autonomous patrolling to identify unexpected situations; in this research, the image matching is done between the observed data and  large databases of normal images.
Recent works use Deep Learning models on images. Christiansen et al.~\cite{christiansen2016deepanomaly} propose DeepAnomaly, a custom CNN derived from AlexNet~\cite{krizhevsky2012imagenet}; the model is used to detect and highlight obstacles or anomalies on an autonomous agricultural robot via high-level features of the CNN layers. 
Wellhausen et el.~\cite{wellhausen2020safe} verify the ground traversability for a legged ANYmal~\cite{hutter2016anymal} robot in unknown environments. The paper compares three models - Deep SVDD~\cite{ruff2018deep}, Real-NVP~\cite{dinh2016density}, and a standard autoencoder - on detecting terrain patches whose appearance is anomalous with respect to the robot's previous experience. All the models are trained on patches of footholds images coming from the robot's previous sorties; the most performing model is the combination of Real-NVP and an encoding network, followed closely by the autoencoder.


\section{Datasets}
\label{sec:dataset}

\begin{figure}[t]
\begin{center}
\begin{small}
\setlength\tabcolsep{2 pt}
\begin{tabular}{cccccc}
   & normal  & dust &  roots & roots & wet \\
\rotatebox[origin=c]{90}{Tunnels}  & \raisebox{\samplerise}{\includegraphics[width=\samplesize]{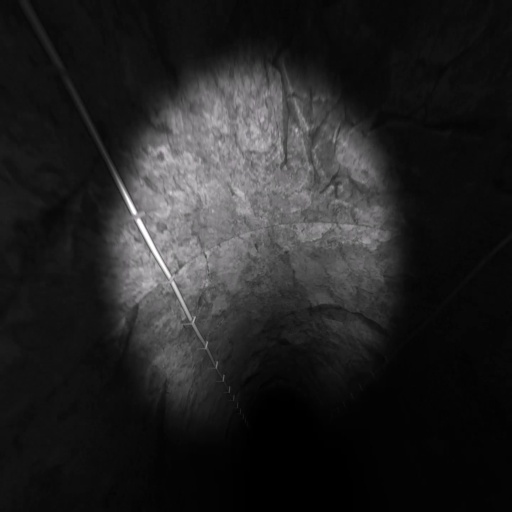}}  &    \raisebox{\samplerise}{\includegraphics[width=\samplesize]{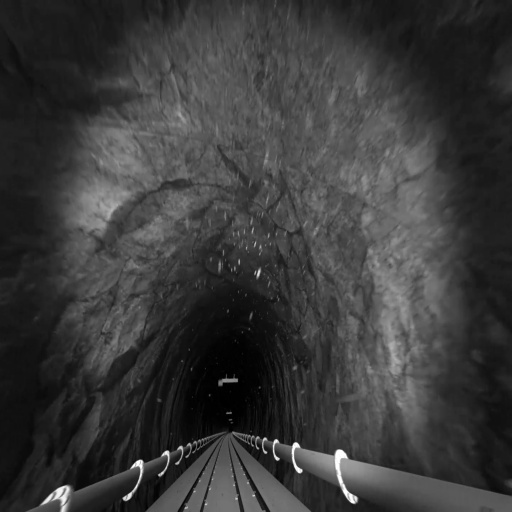}}  &
                \raisebox{\samplerise}{\includegraphics[width=\samplesize]{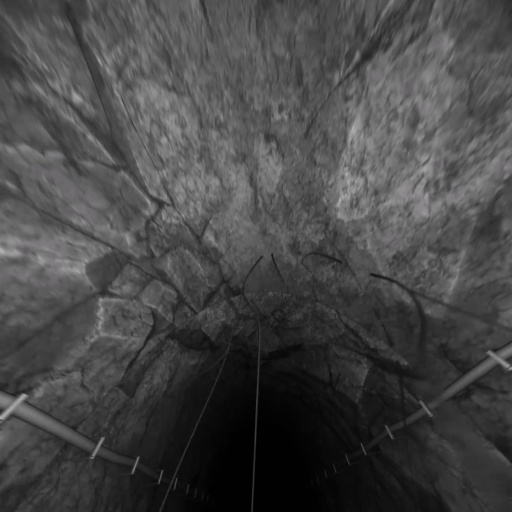}}   &   \raisebox{\samplerise}{\includegraphics[width=\samplesize]{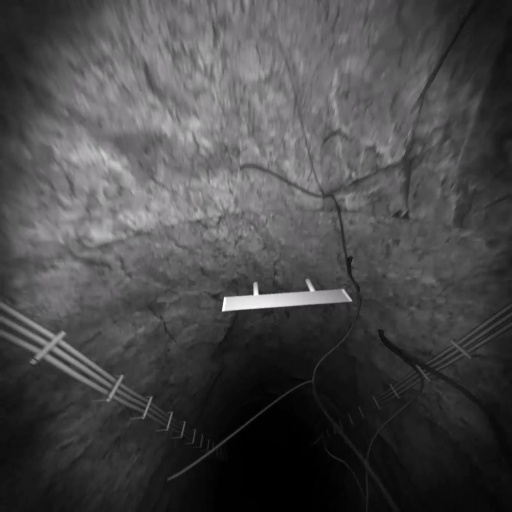}}   &   \raisebox{\samplerise}{\includegraphics[width=\samplesize]{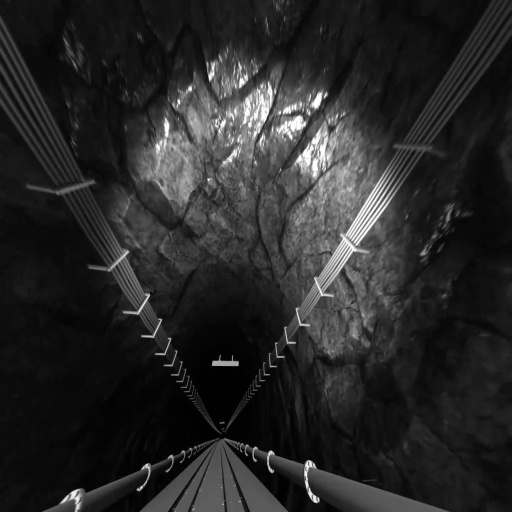}}  \\
 & normal   & mist &  mist & tape & tape \\
\rotatebox[origin=c]{90}{Factory}  &  \raisebox{\samplerise}{\includegraphics[width=\samplesize]{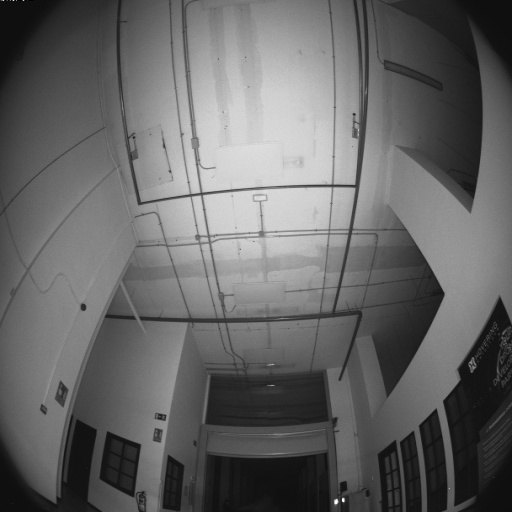}}   &    \raisebox{\samplerise}{\includegraphics[width=\samplesize]{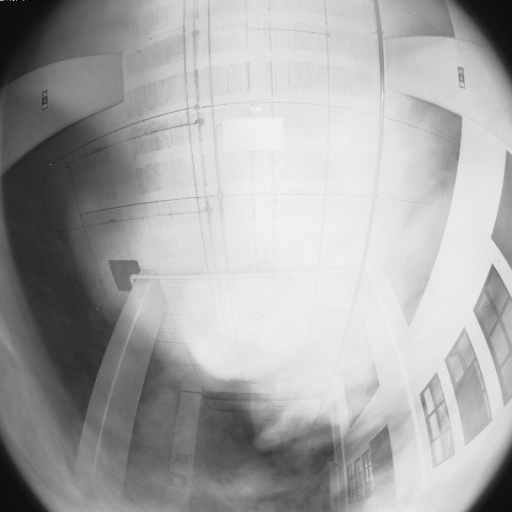}}  &
                \raisebox{\samplerise}{\includegraphics[width=\samplesize]{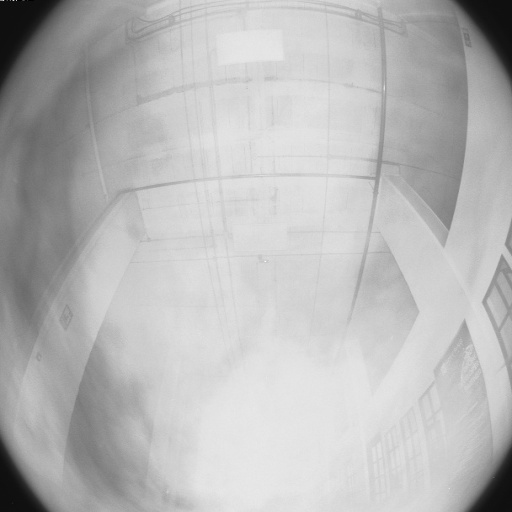}}   &   \raisebox{\samplerise}{\includegraphics[width=\samplesize]{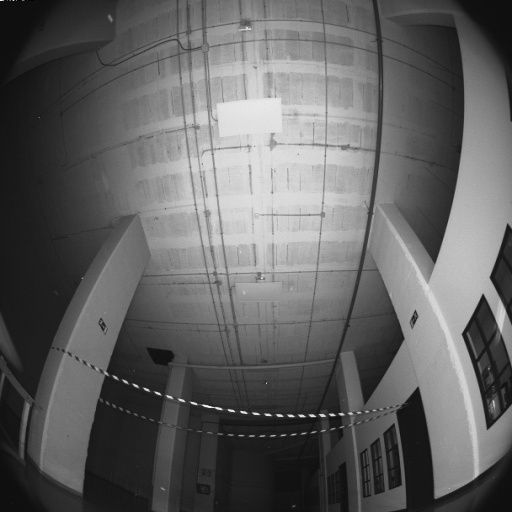}}   &   \raisebox{\samplerise}{\includegraphics[width=\samplesize]{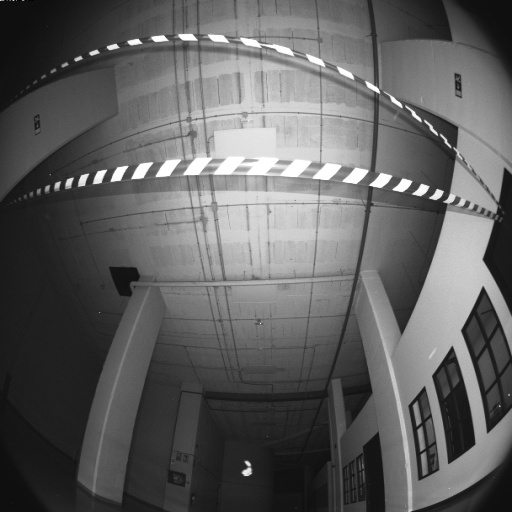}}  \\
 & normal  & human &  floor & screws & defect \\
\rotatebox[origin=c]{90}{Corridors}   &  \raisebox{\samplerise}{\includegraphics[width=\samplesize]{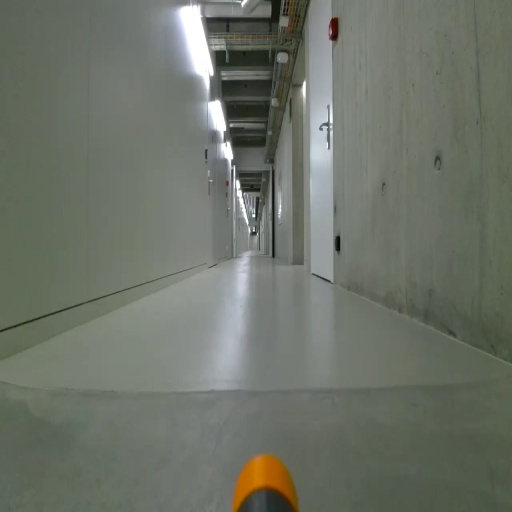}}    &   \raisebox{\samplerise}{\includegraphics[width=\samplesize]{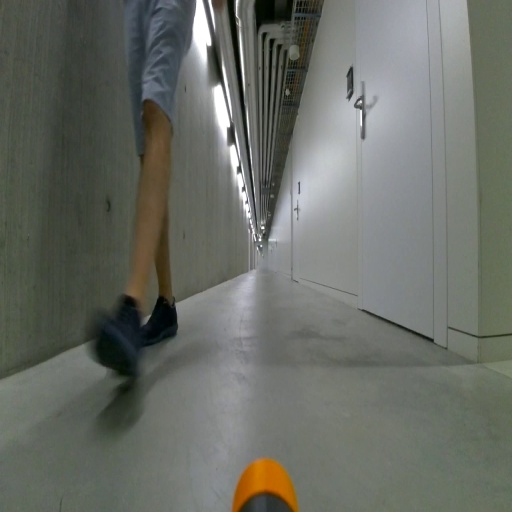}}  &
                \raisebox{\samplerise}{\includegraphics[width=\samplesize]{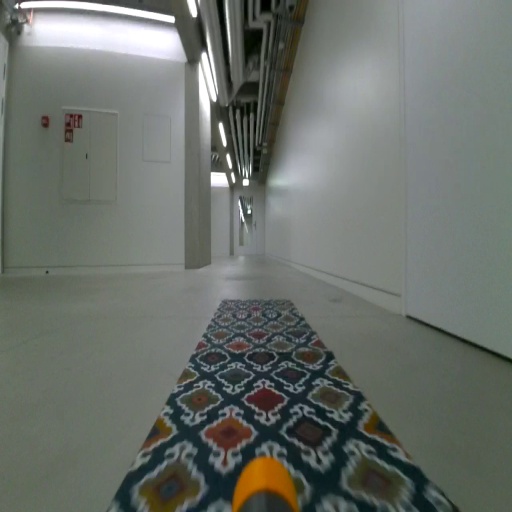}}   &   \raisebox{\samplerise}{\includegraphics[width=\samplesize]{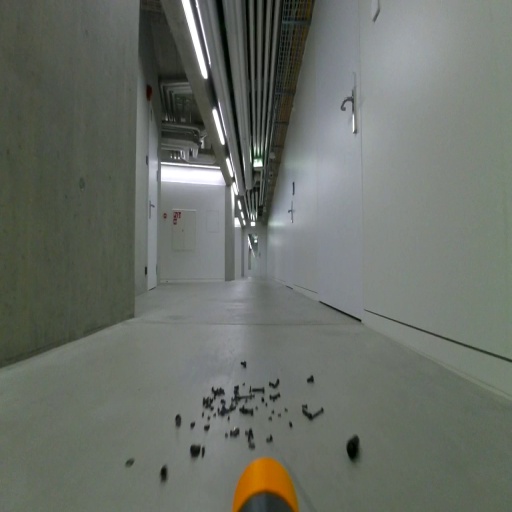}}   &   \raisebox{\samplerise}{\includegraphics[width=\samplesize]{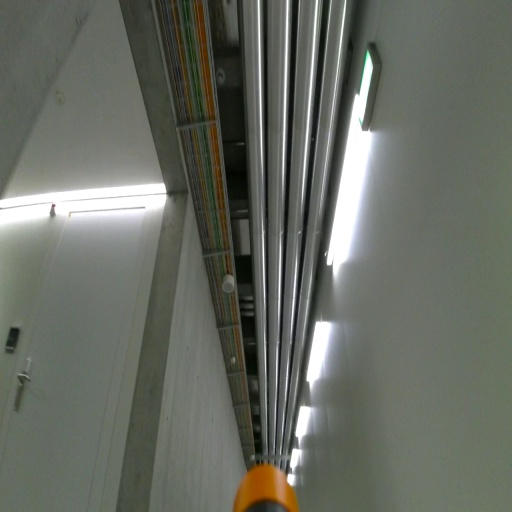}}  \\
   & normal & cable & cellophane & hang. cable &   water  \\
\rotatebox[origin=c]{90}{Corridors}  & \raisebox{\samplerise}{\includegraphics[width=\samplesize]{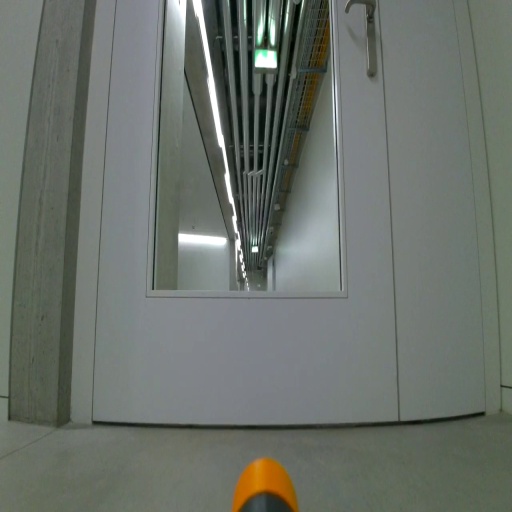}}  &  \raisebox{\samplerise}{\includegraphics[width=\samplesize]{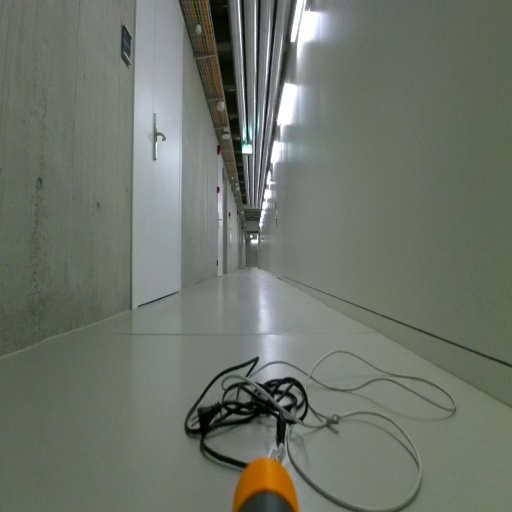}}   &   \raisebox{\samplerise}{\includegraphics[width=\samplesize]{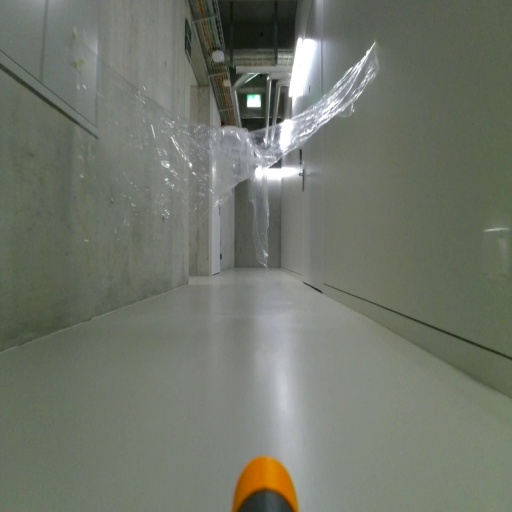}}   &   \raisebox{\samplerise}{\includegraphics[width=\samplesize]{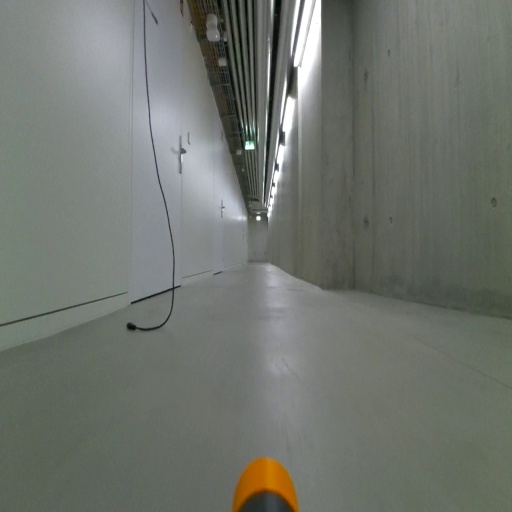}} &
                \raisebox{\samplerise}{\includegraphics[width=\samplesize]{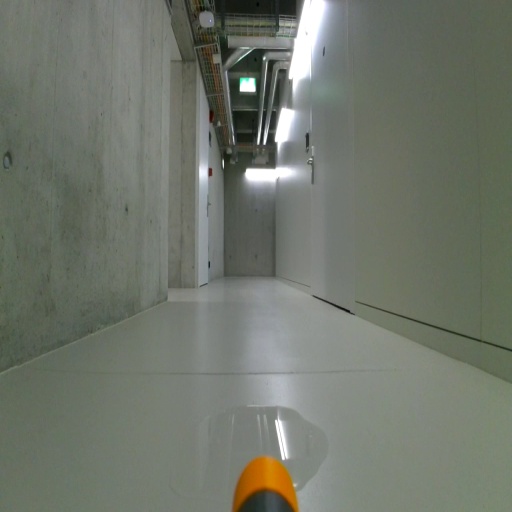}}   \\
\end{tabular}
\end{small}
\end{center}
\caption{The testing datasets are composed of normal images and by images of different anomaly classes.}
\label{fig:frames}
\end{figure}

We contribute three datasets representing different operating scenarios for indoor robots (flying or ground).  Each dataset is composed of a large number of grayscale or RGB frames with a $512 \times 512$ px resolution. For each dataset, we define four subsets:
\begin{itemize}
    \item a training set, composed of only normal frames;
    \item a validation set, composed of only normal frames;
    \item a labeled testing set, composed of frames with an associated label; some frames in the testing set are normal, others are anomalies and are associated with the respective anomaly class;
    \item an unlabeled qualitative testing set, consisting of one or more continuous data sequences acquired at approximately 30 Hz, depicting the traversal of environments with alternating normal and anomalous situations.
\end{itemize}
The training set is used for training anomaly detection models, and the validation set for performing model-selection of reconstruction-based approaches -- e.g., determining when to stop training an autoencoder.
The testing set can be used to compute quantitative performance metrics for the anomaly detection problem.  The qualitative testing set can be used to analyze how the model, the autoencoder in our case, outputs react to a video stream as the robot traverses normal and anomalous environments.

The very concept of anomaly in robotic perception is highly subjective and application-dependent~\cite{birnbaum2015unmanned,christiansen2016deepanomaly,wellhausen2020safe}.  Whether a given situation should be considered an anomaly depends on the features of the robot and on its task;  for example, consider a robot patrolling corridors with floors normally clear of objects; the presence of screws and bolts littering the ground could be hazardous for a robot with inflated tires that could get punctured, but completely irrelevant for a drone or legged robot.  On an orthogonal dimension, some applications might be interested in determining anomalies regardless of whether they pose a hazard to the robot: in a scenario in which a robot is patrolling normally-empty tunnels, finding anything different in the environment could be a sign of an intrusion and should be detected.  The appearance of anomalies in forward-looking camera streams is also dependent on the distance from the robot; wires or other thin objects that might pose a danger to a robot could be simply invisible if they are not very close to the camera.  Our labeled testing sets are manually curated, and we used our best judgment to determine whether to consider a frame anomalous or not: frames with anomalies that are not clearly visible in the $512\times512$ full-resolution images are excluded from the quantitative testing set, but they are preserved in the qualitative testing sequences.


\subsection{\emph{Tunnels} Dataset}
\label{sec:tunnels}
The dataset, provided by Hovering Solutions Ltd, is composed of grayscale frames from simulated drone flights along procedurally-generated underground tunnels presenting features typically found in aqueduct systems, namely: random dimensions; random curvature radius; different structures on the floor; tubing, wiring, and other facilities attached to the tunnel walls at random positions; uneven textured walls; various ceiling-mounted features at regular intervals (lighting fixtures, signage).  The drone flies approximately along the centerline of the tunnel and illuminates the tunnel walls with a spotlight approximately coaxial with the camera. Both the camera and the spotlight are slightly tilted upwards.  

This dataset is composed of $143070$ frames: $72854$ in the training set; $8934$ in the validation set; $57081$ in the quantitative labeled testing set ($40\%$ anomalous); $4201$ in the qualitative testing sequences.

Three anomalies are represented: \emph{dust}, \emph{wet} ceilings, and thin plant \emph{roots} hanging from the ceilings (see Figure~\ref{fig:frames}).  These all correspond to hazards for quadrotors flying real-world missions in aqueduct systems: excessive amounts of dust raised by rotor downwash hinder visual state estimation; wet ceilings, caused by condensation on cold walls in humid environments, indicate the risk of drops of water falling on the robot; thin hanging roots, which find their way through small cracks in the ceiling, directly cause crashes.

\subsection{\emph{Factory} Dataset}
\label{sec:factory}
This dataset contains grayscale frames recorded by a real drone, with a similar setup to the one simulated in the Tunnels dataset, flown in a testing facility (a factory environment) at Hovering Solutions Ltd. During acquisition, the environment is almost exclusively lit by the onboard spotlight.

This dataset is composed of $12040$ frames: $4816$ in the training set; $670$ in the validation set; $6001$ in the quantitative testing set ($53\%$ anomalous); $553$ in the qualitative testing sequences.

Two anomalies are represented: \emph{mist} in the environment, generated with a fog machine; and a signaling \emph{tape} stretched between two opposing walls (Figure~\ref{fig:frames}). These anomalies represent large-scale and small-scale anomalies, respectively.

\subsection{\emph{Corridors} Dataset}
\label{sec:corridors}
This dataset contains RGB frames recorded by a real teleoperated omni-directional ground robot (DJI Robomaster S1), equipped with a forward-looking camera mounted at \SI{22.5}{cm} from the ground, as it explores corridors of the underground service floor of a large university building. The corridors have a mostly uniform, partially reflective floor with few features; various side openings of different size (doors, lifts, other connecting corridors); variable features on the ceiling, including service ducts, wiring, and various configurations of lighting.  The robot is remotely teleoperated during data collection, traveling approximately along the center of the corridor.

This dataset is composed of $52607$ frames: $25844$ in the training set; $2040$ in the validation set; $17971$ in the testing set ($45\%$ anomalous); $6752$ in qualitative testing sequences.

8 anomalies are represented, ranging from subtle characteristics of the environment affecting a minimal part of the input to large-scale changes in the whole image acquired by the robot: \emph{water} puddles, \emph{cables} on the floor; \emph{hanging cables} from the ceiling; different mats on the \emph{floor}, \emph{human} presence, \emph{screws} and bolts on the ground; camera \emph{defects} (extreme tilting, dirty lens) and \emph{cellophane} foil stretched between the walls.  Examples of these anomalies are in Figure~\ref{fig:frames}. 

\section{Experimental Setup}
\label{sec:experimental_setup}

\subsection{Anomaly Detection on Frames}
\begin{figure}[t]
\begin{center}
\includegraphics[width=1\textwidth]{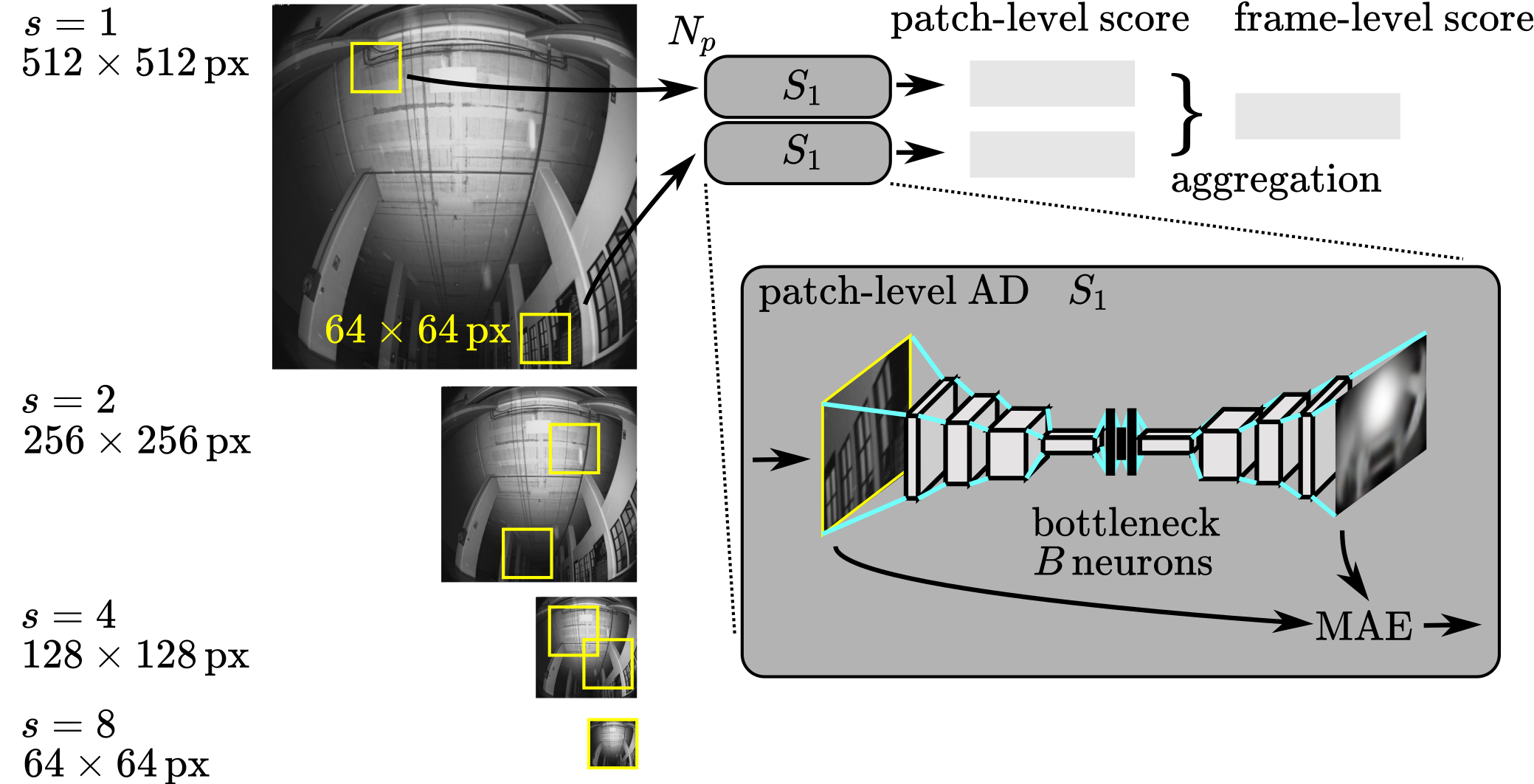}
\caption{Anomaly detection model: using an autoencoder to compute the patch-level anomaly scores, which are aggregated in a frame-level score.}
\label{fig:anomaly-model}
\end{center}
\end{figure}

We define an anomaly detector as a function mapping a frame ($512 \times 512$) to an anomaly score, which should be high for anomalous frames and low for normal ones.  The frame-level anomaly detector relies on a patch-level anomaly detector (see Figure~\ref{fig:anomaly-model}), which instead operates on low-resolution inputs ($64 \times 64$), which is a typical input size for anomaly detection methods operating on images~\cite{wellhausen2020safe,kerner2019novelty}.

First, the frame is downsampled (using local averaging) by a factor $s \in \{1,2,4,8\}$; we will refer to the respective models as $S_{1}$, $S_{2}$, $S_{4}$ and $S_{8}$.  The resulting downsampled image, with resolution $512/s \times 512/s$, is standardized to zero mean and unit variance, independently for each channel; we then extract $N_p$ $64 \times 64$ patches, at random coordinates, such that they are fully contained in the downsampled image.
The patch-level anomaly detector is applied to each patch, producing $N_p$ anomaly scores; these are aggregated together (e.g., computing their average) to yield the frame-level anomaly score.

Note that in the case of $S_{8}$, $N_p\equiv1$ since a unique patch can be defined on a $64\times64$ downsampled image.  This corresponds to the special case in which the whole frame (after downsampling) is directly used as input to the patch-based detector.  This approach is simple and attractive but is unsuitable to detect small-scale anomalies since it can not leverage the full resolution of the frame.



\subsection{Patch-Level Anomaly Detector}
\label{sec:autoencoder}

Patch-level anomalies are detected with a standard approach based on the reconstruction error of an autoencoder.
The encoder part operates on a $64 \times 64$ input and is composed of four convolutional layers with a LeakyReLU activation function; each layer has a number of filters that is double the number of filters of the previous layer; we start with $F$ $3\times3$ filters for the first layer. Each Convolution has stride 2 thus halving the resolution of the input. The neurons of the last layer of the encoder are flattened and used as input to a fully connected layer with $B$ neurons (bottleneck); the decoder is built in a specular manner to the encoder, and its output has the same shape as the encoder's input; the output layer has a linear activation function, which enables the model to reconstruct the same range as the input.
During inference, the patch-based anomaly detector accepts a patch as input and outputs the Mean Absolute Error between the input patch and its reconstruction, which we interpret as the patch anomaly score.

\subsection{Training}
For a given scale $s$, the autoencoder is trained as follows: first, we downsample each frame in the training set by a factor $s$; then, as an online data generation step, we sample random patches completely contained in the downsampled frames.

We use the Adam~\cite{kingma2014adam} optimizer to minimize the mean squared reconstruction error, with an initial learning rate of $0.001$, which is reduced by a factor of $10$ in case the validation loss plateaus for more than 8 epochs.  Because the size of the training set of different datasets is widely variable, we set the total number of epochs in such a way that during the whole training, the model sees a total of $2$ million samples; this allows us to better compare results on different datasets. 

The approach is implemented in PyTorch and Python 3.8, using a deep learning workstation equipped with 4 NVIDIA 2080 Ti GPUs; training each model takes about $\SI{1}{\hour}$ on a single GPU.


\begin{table*}[t!]
\caption{AUC values for models at all scales}
\begin{center}
\scriptsize
\begin{tabular}{l|r|rrrr|rrr|rrrrrrrrr}
\toprule
scale & \textbf{Avg} & \multicolumn{4}{c}{Tunnels} & \multicolumn{3}{c}{Factory} & \multicolumn{9}{c}{Corridors} \\
      &     \rot{\textbf{all}} &    \rot{\textbf{all}} & \rot{dust} & \rot{root} &  \rot{wet} &     \rot{\textbf{all}} & \rot{mist} & \rot{tape} &      \rot{\textbf{all}} & \rot{water} & \rot{cellophane} & \rot{cable} & \rot{defect} & \rot{hang. cable} & \rot{floor} & \rot{human} & \rot{screws} \\
\midrule
   $S_8$ & \textbf{0.82} &   \textbf{0.82} & 0.54 & 0.76 & 0.87 &    \textbf{0.90} & 0.95 & 0.48 &     \textbf{0.74} &  0.63 &       0.66 &  0.70 &    1.00 &          0.44 &  0.85 &  0.67 &   0.48 \\
   $S_4$ & \textbf{0.62} &   \textbf{0.89} & 0.62 & 0.79 & 0.94 &    \textbf{0.24} & 0.25 & 0.17 &    \textbf{0.73} &  0.81 &       0.70 &  0.81 &    1.00 &          0.41 &  0.38 &  0.73 &   0.30 \\
   $S_2$ & \textbf{0.60} &   \textbf{0.88} & 0.63 & 0.80 & 0.93 &    \textbf{0.21} & 0.20 & 0.30 &    \textbf{0.71} &  0.78 &       0.70 &  0.75 &    0.99 &          0.50 &  0.51 &  0.56 &   0.40 \\
   $S_1$ & \textbf{0.55} &   \textbf{0.85} & 0.61 & 0.80 & 0.88 &    \textbf{0.12} & 0.10 & 0.25 &    \textbf{0.69} &  0.72 &       0.73 &  0.68 &    0.90 &          0.60 &  0.59 &  0.51 &   0.55 \\
\bottomrule
\end{tabular}
\end{center}
\label{table:results}
\end{table*}

\subsection{Metrics}
\label{sec:metric}

We evaluate the performance of the frame-level anomaly detector on the testing set of each dataset.  In particular, we quantify the anomaly detection performance as if it was a binary classification problem (normal vs anomalous), where the probability assigned to the anomalous class corresponds to the anomaly score returned by the detector.  This allows us to define the Area Under the ROC Curve metric (AUC); an ideal anomaly detector returns anomaly scores such that there exists a threshold $t$ for which all anomalous frames have scores higher than $t$, whereas all normal frames have scores lower than $t$: this corresponds to an AUC of 1.  An anomaly detector returning a random score for each instance, or the same score for all instances, yields an AUC of 0.5.  The AUC value can be interpreted as the probability that a random anomalous frame is assigned an anomaly score larger than that of a random normal frame.  The AUC value is a meaningful measure of a model's performance and does not depend on the choice of threshold.

For each model and dataset, we compute the AUC value conflating all anomalies, as well as the AUC individually for each anomaly (versus normal frames, ignoring all other anomalies).


\section{Results}

\label{sec:experimental_results}

\subsection{$S_8$ Model Hyperparameters}
\label{sec:results_model}

\begin{figure}[h!]
\begin{minipage}{\textwidth}
\centering
\includegraphics[height=0.6cm]{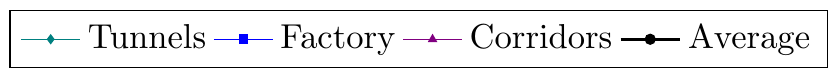}
\end{minipage}
\begin{minipage}{0.48\textwidth}
\subfloat[Results for different autoencoder's bottleneck sizes for model $S_8$. Top two rows: for the same two samples  (normal in green, anomalous in red), autoencoder reconstructions. Center: score distributions over the testing set. Bottom: mean score difference between anomalous and normal samples and AUC of the anomaly detector.\label{fig:results_bottleck}]{
\includegraphics[width=\textwidth]{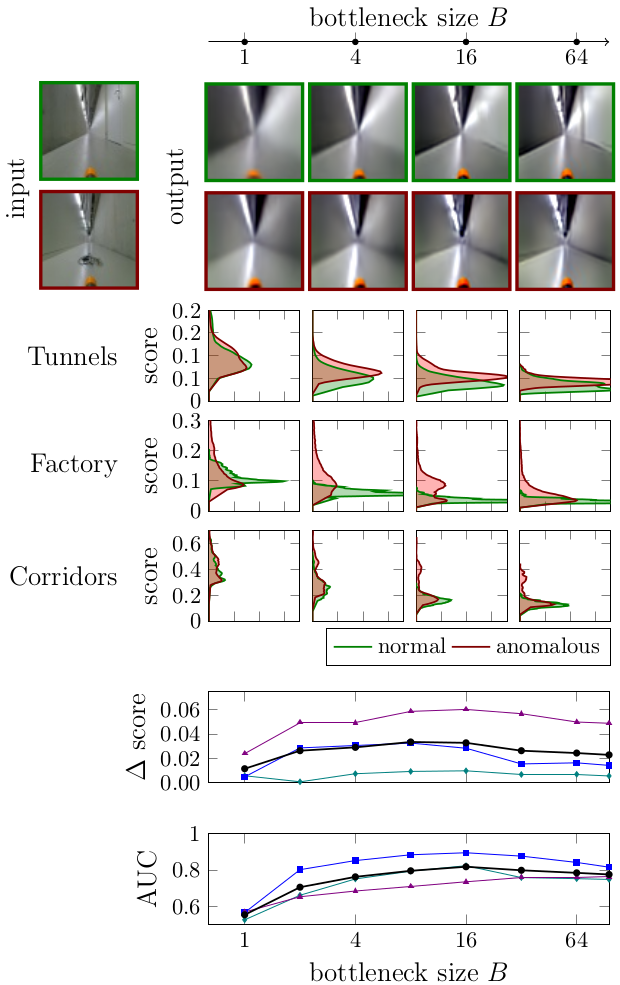}
}
\end{minipage}
\hfill
\begin{minipage}{0.48\textwidth}
\vspace{0.8mm}
\subfloat[Results for model $S_{8}$ for autoencoders with different first layer sizes.\label{fig:results_first_layer}]{\includegraphics[width=\textwidth]{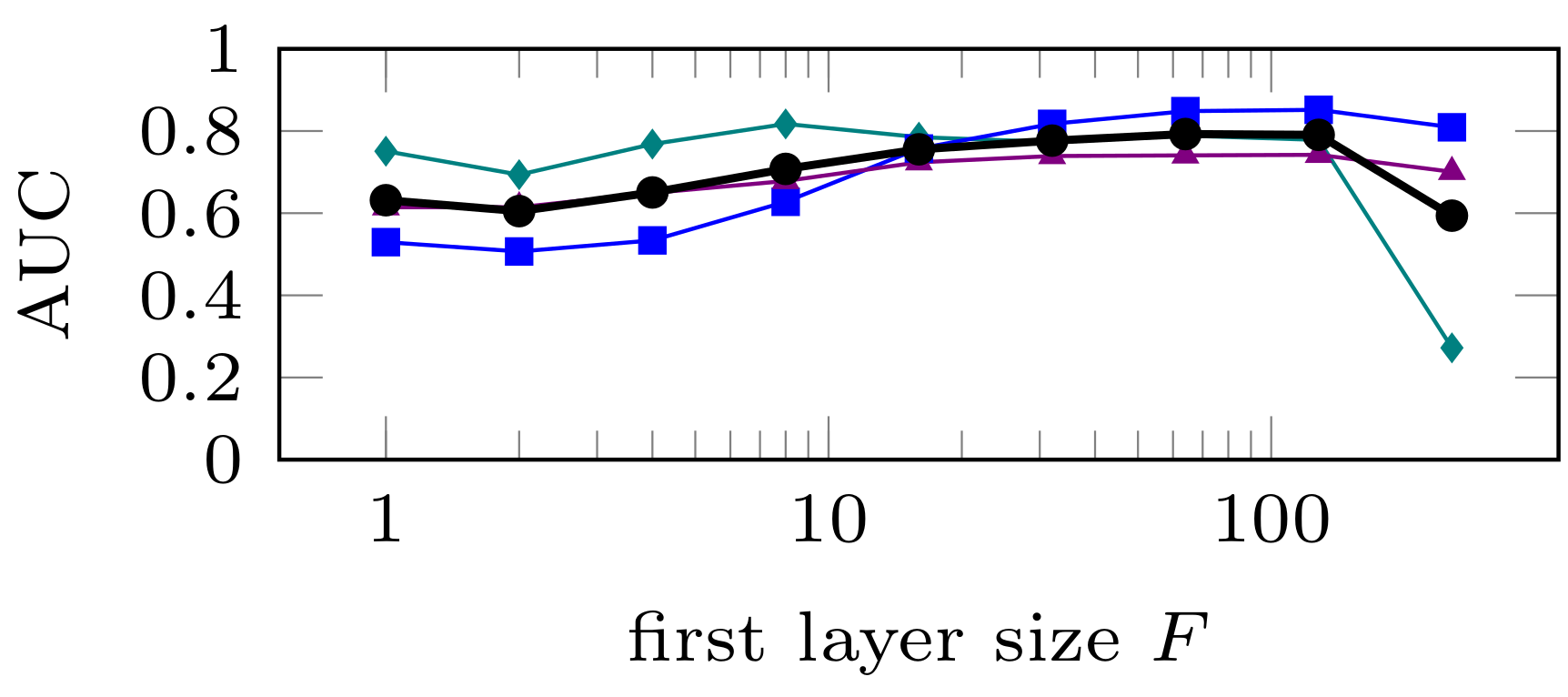}}\\
\subfloat[Results for model $S_{2}$ when aggregating the scores of multiple patches extracted from each frame. Top: AUC, when aggregating by averaging, for different numbers of patches, compared to $S_8$ (dotted). Bottom: AUC, when aggregating 250 patches by computing a quantile (solid) or by averaging (dashed).\label{fig:results_aggregation}
]{\includegraphics[width=\textwidth]{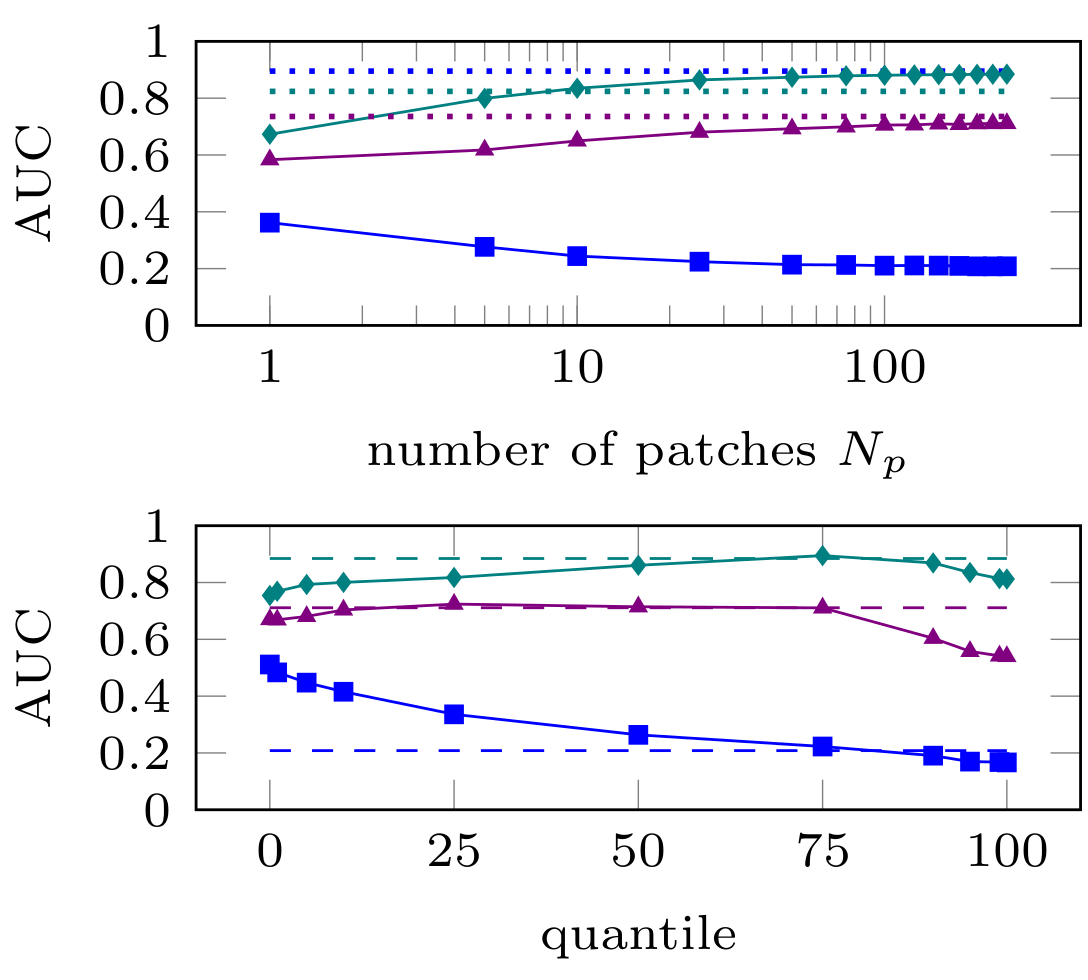}}\\
\end{minipage}
\caption{Experimental results}
\end{figure}

Figure~\ref{fig:results_bottleck} explores the choice of the bottleneck size $B$ for model $S_8$.  Increasing $B$ reduces reconstruction error for both anomalous and normal data; the reconstruction error best discriminates the two classes (higher AUC, higher average gap between the two classes) for intermediate values of $B$ ($16$ neurons): then, the autoencoder can reconstruct well normal data while lacking the capacity to properly reconstruct anomalous samples.  These findings apply to all three datasets.
%
%
Figure~\ref{fig:results_first_layer} investigates a similar capacity trade-off: autoencoders with a small number of filters for the first convolution layer (\emph{first layer size}) are not powerful enough to reproduce well even normal samples, therefore have lower discriminating performance.
For the rest of the Section, we only report results for bottleneck size $B=16$ and first layer size $F=128$.

\subsection{Patch Aggregation}
\label{sec:aggregation-results}

Figure~\ref{fig:results_aggregation}:top explores the impact of $N_p$ on the anomaly detection performance of model $S_2$; we observe that, for the Tunnels and Corridors datasets, the performance increases as $N_p$ increases.  This is expected, as more patches are processed and aggregated to compute the frame-level score. 
Only for Tunnels, $S_2$ outperforms $S_8$ for 10 or more patches.

On the contrary, for the Factory dataset, the model $S_2$ performs worse than chance at detecting anomalies and assigns lower scores than normal data. 
this is due to the testing set being dominated by the mist anomaly, which is not detectable at low scales as discussed in Section~\ref{sec:scale_and_anomalies}.

Figure~\ref{fig:results_aggregation}:bottom reports how computing the 0.7-0.8 quantile offers a slightly better aggregation than averaging.

\subsection{Scales and Anomalies}
\label{sec:scale_and_anomalies}

\begin{figure}[htb]
\begin{center}
\includegraphics[width=\textwidth]{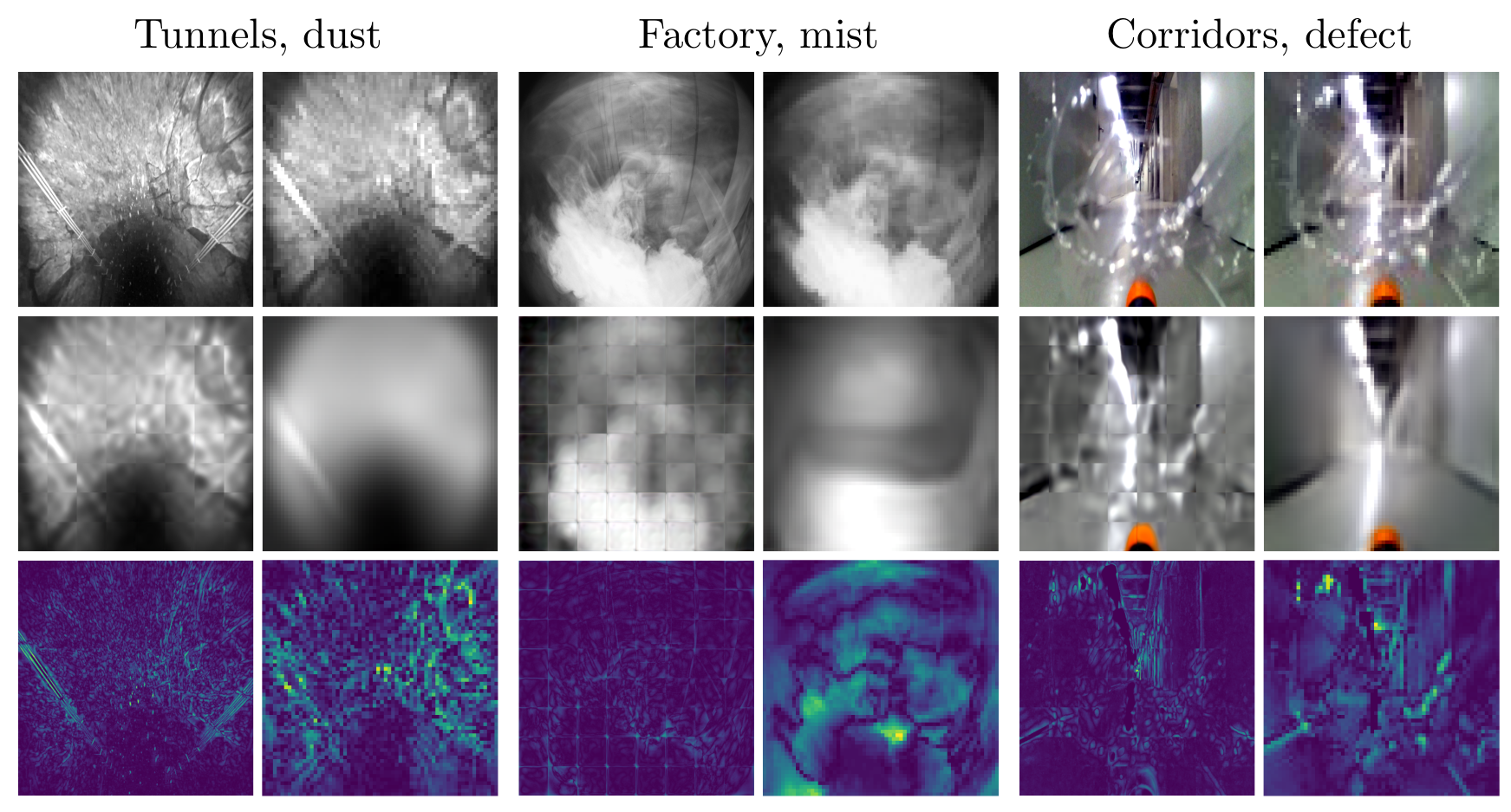}
\caption{Comparison between $S_1$ and $S_8$: pairs of identical input images representing an anomaly (top row); Autoencoder's outputs (central row) for $S_1$ (left) and $S_8$ (right); the absolute value of the difference between input and output (bottom row, using a colormap where yellow is high and blue is low).
Only for this illustration, for $S_1$ we regularly sample 64 patches to cover the whole input image, and we use the output patches to compose a full-resolution image.
}
\label{fig:reconstructions}
\end{center}
\end{figure}

Table~\ref{table:results} summarizes the most important results on all model scales, datasets, and anomalies.  We note that most anomalies are best detected by the full-frame approach $S_{8}$; this is especially true for large-scale anomalies that cover a significant portion of the frame, such as mist for Factory, or human and floor for Corridors.  In contrast, $S_8$ underperforms for small-scale anomalies, that cover few pixels of the downsampled image (e.g., dust and roots for Tunnels; cellophane, water, and hanging cable for Corridors); in this case, small-scale models sometimes have an advantage over $S_8$.

In contrast, we observe that small-scale models struggle with the large-scale mist anomaly, returning consistently lower anomaly scores than normal frames, which yields AUC values well below 0.5.  Figure~\ref{fig:reconstructions} compares how $S_1$ and $S_8$ reconstruct a mist frame: clearly, $S_8$ fails to capture the large-scale structure of mist, which yields high reconstruction error as expected in an anomalous frame; in contrast, since individual high-resolution patches of the mist frame are low-contrast and thus easy to reconstruct, the $S_1$ model yields very low reconstruction error and, thus, low AUC.

Some anomalies, such as defect for Corridors, are obvious enough that models at all scales can detect them almost perfectly.



\subsection{Run-time Evaluation}
\label{sec:runtime}

\begin{figure}[ht]
\begin{center}
\includegraphics[width=0.9\textwidth]{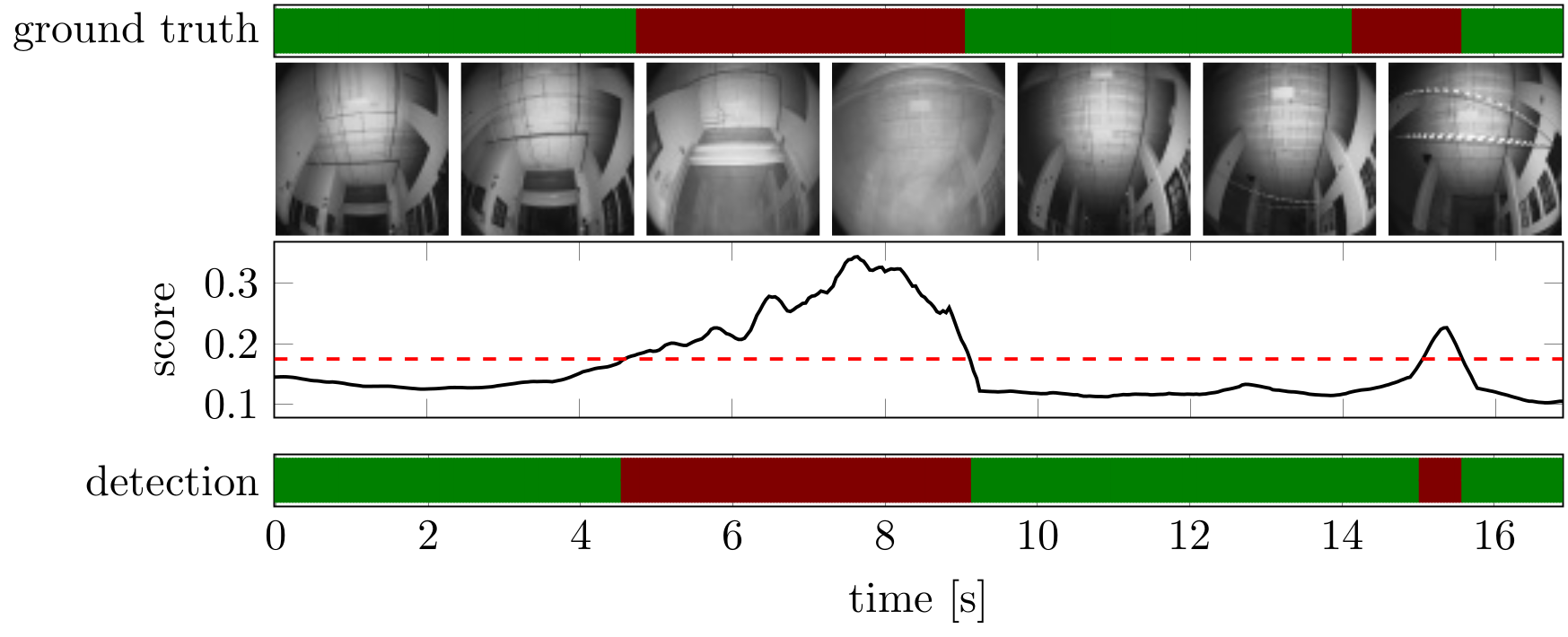}
\caption{Part of the qualitative testing dataset in the Factory scenario where the drone first passes through mist and then below a tape. Top: the manually added labels (green: normal, red: anomalous) and seven frames sampled from the timeline. Center: the score returned by the $S_{8}$ model (solid black) and anomaly threshold $t$ (dashed red).
Bottom: the anomaly detector output. 
}
\label{fig:results_validation}
\end{center}
\end{figure}

The accompanying video features several runs where a quadcopter uses the $S_8$ model to detect anomalies on-board to recognize and avoid unforeseen hazards.  Figure~\ref{fig:results_validation} illustrates execution on a sequence that is part of the qualitative testing set for Factory; in the figure, we manually annotated the ground truth presence of hazards such as mist (first red interval) and tape (second red interval).  In the experiment, the robot captures a camera frame, computes an anomaly score, and raises an alarm when the score passes a predefined threshold. The example shows how the drone is able to detect first a long area of mist and later a small signaling tape.

\section{Conclusions}
We introduced three datasets for validating approaches to detect anomalies in visual sensing data acquired by mobile robots exploring an environment; various anomalies are represented, spanning from camera malfunctions to environmental hazards: some affect the acquired image globally; others only impact a small portion of it.  We used these datasets to benchmark an anomaly detection approach based on autoencoders operating at different scales on the input frames.  Results show that the approach is successful at detecting most anomalies (detection performance with an average AUC metric of 0.82); detecting small anomalies is in general harder than detecting anomalies that affect the whole image.
%

\bibliographystyle{IEEEtran}
\bibliography{biblio}




\end{document}